\documentclass[sigconf, nonacm, authorversion]{acmart}
\AtBeginDocument{%
  \providecommand\BibTeX{{%
    \normalfont B\kern-0.5em{\scshape i\kern-0.25em b}\kern-0.8em\TeX}}}

\begin{document}

\title{Using a Language Model in a Kiosk Recommender System at Fast-Food Restaurants}

\author{Eduard Zubchuk}
\email{ezubchuk@ntr.ai}
\orcid{0000-0002-0931-6808}
\affiliation{%
  \institution{Higher IT School of Tomsk State University \& NTR Labs}
  %%\streetaddress{P.O. Box 1212}
  \city{Tomsk}
  %%\state{Ohio}
  \country{Russia}
  \postcode{634050}
}

\author{Dmitry Menshikov}
\email{menshikov.dmitry@gmail.com}
\affiliation{%
  \institution{Higher IT School of Tomsk State University \& NTR Labs}
  \city{Moscow}
  \country{Russia}
}

\author{Nikolay Mikhaylovskiy}
\email{nickm@ntr.ai}
\orcid{0000-0001-5660-0601}
\affiliation{%
  \institution{Higher IT School of Tomsk State University \& NTR Labs}
  \city{Moscow}
  \country{Russia}
}

%%
%% By default, the full list of authors will be used in the 
%% \renewcommand{\shortauthors}{Trovato and Tobin, et al.}

\begin{abstract}
  Kiosks are a popular self-service option in many fast-food restaurants, they save time for the visitors and save labor for the fast-food chains. In this paper, we propose an effective design of a kiosk shopping cart recommender system that combines a language model as a vectorizer and a neural network-based classifier. The model performs better than other models in offline tests and exhibits performance comparable to the best models in A/B/C tests. 
\end{abstract}



  %%\includegraphics[width=\textwidth]{sampleteaser}
  %%\caption{Seattle Mariners at Spring Training, 2010.}
  %%\Description{Enjoying the baseball game from the third-base
  %%seats. Ichiro Suzuki preparing to bat.}
  %%\label{fig:teaser}

\maketitle

\section{Introduction}
Kiosks are a popular self-service option in many fast-food restaurants, they save time for the customers and save labor for the fast-food chains. With the advent of COVID-19 pandemic, minimizing in-person interaction drives faster adoption of kiosks by fast-food chains. A recommender system for kiosks should allow increasing revenue per visitor, by creating a unique user experience whereby:
\begin{itemize}
\item the fast-food restaurant visitor would be regularly exposed to the recommendations;
\item recommendations will stimulate the purchase;
\item visitors' loyalty will not degrade due to intrusiveness.
\end{itemize}
In this work, we describe the design of one of the pilot recommender systems for kiosks of a fast-food chain, a shopping cart recommender system. The goal of this recommender system is to recommend items based on the interactions of the visitor with the kiosk during the session, specifically just before checkout. The systems piloted were compared using A/B/C tests measuring the gross margin gained from the items sold by recommendation. Thus the navigational function of a recommender system was out of consideration in this test. The validity of this measurement approach was supported by the fact that virtually no visitors returned from the shopping cart to general items selection. The system was piloted in 100+ fast-food restaurants for a prolonged period of time.
\subsection{Our contribution}
The contribution of our work is twofold. First, we propose a novel architecture for a recommender system. The architecture consists of a vectorizer that turns a shopping cart into a vector, and a classifier of such vectors, trained separately. Second, we show that using the FastText model  \cite{bojanowski2017enriching} as a vectorizer and a fully connected neural network (Multi-Layer Perceptron) as a classifier delivers competitive results for a fast-food kiosk shopping cart recommender system.
\section{Problem Statement}
\begin{figure}
  \includegraphics[width=0.4\textwidth]{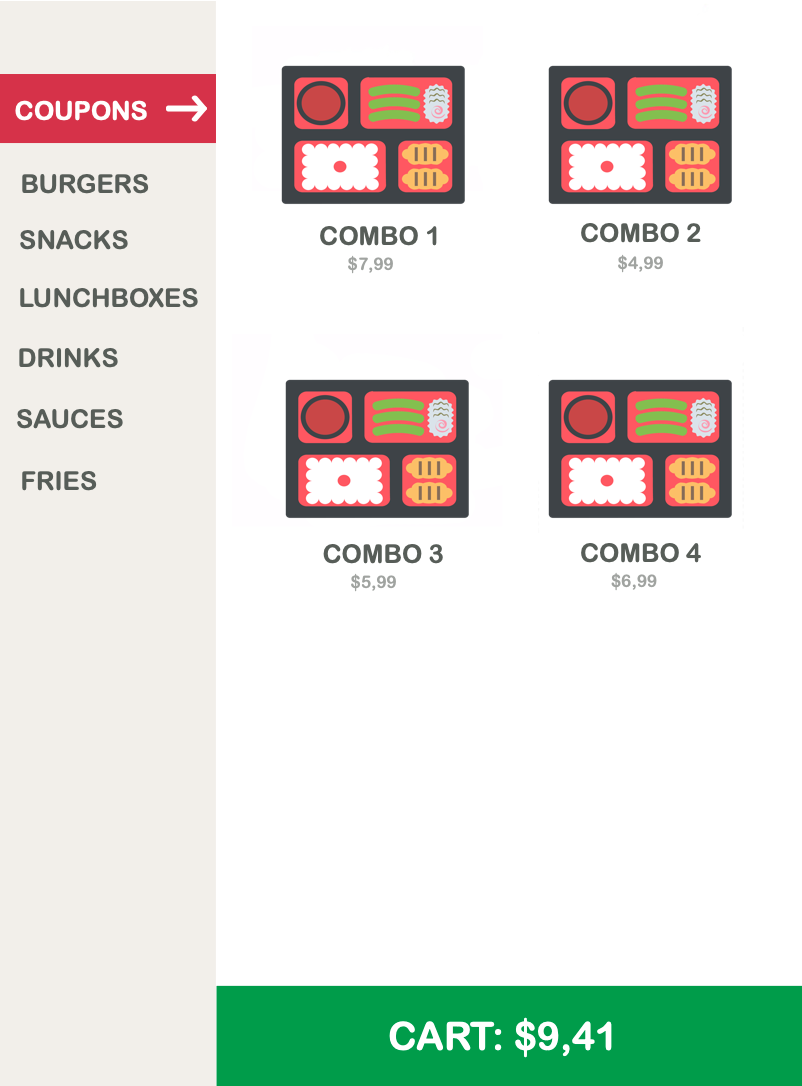}
  \caption{General layout of the kiosk interface.}
   \label{fig1}
\end{figure}

\begin{figure}
  \includegraphics[width=0.4\textwidth]{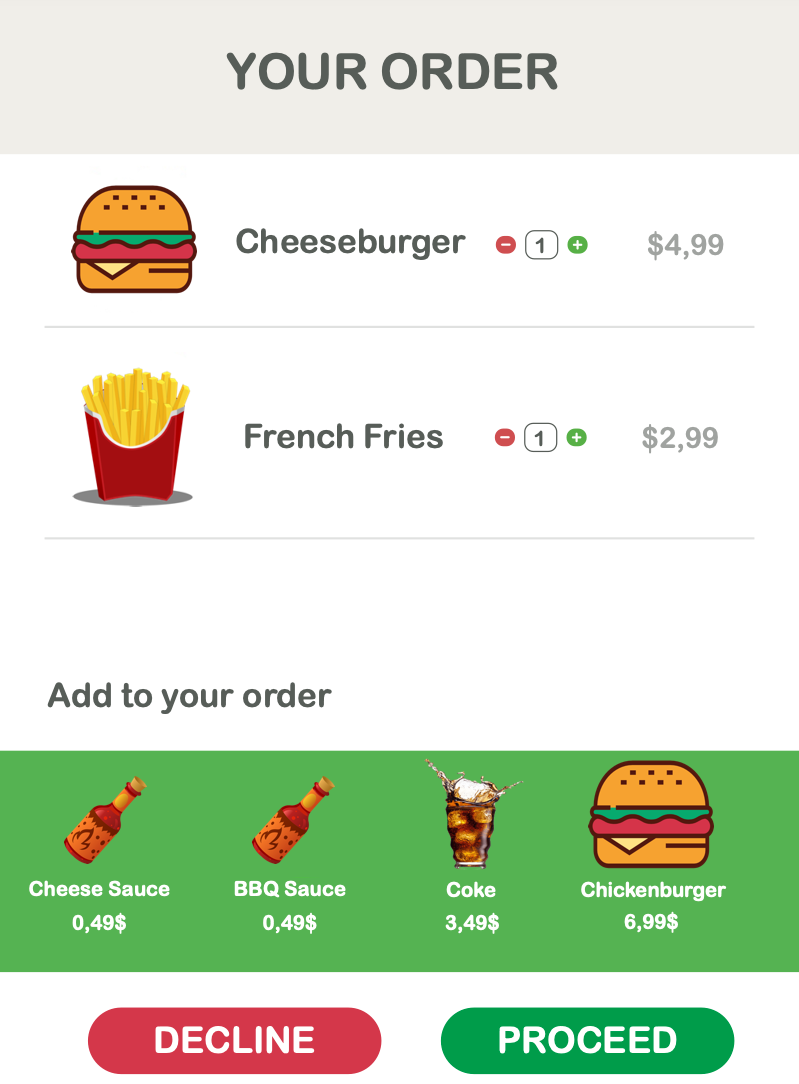}
  \caption{General layout of the shopping cart}
   \label{fig2}
\end{figure}

The order placement process in a self-service kiosk in a fast-food restaurant usually includes at least menu browsing and checkout. During the browsing phase (see Figure~\ref{fig1}), the visitor adds items of interest to the shopping cart, while the checkout is aimed for validation of the order and payment. There are a few options for placing the recommendations during the purchase process. Our effort was focused on recommendations at the checkout phase.  Figure~\ref{fig2} represents the shopping cart screen. Green (best viewed on screen) section under the line “Add to your order” is a recommendation section that includes four items from the main menu. Users can add these items before proceeding to payment. It is important to note that in this layout the order of the recommended items is not important; separate A/B/C tests have shown that there is no statistically significant differences induced by the order of the items recommended.

Our task was to recommend four items from the menu based on the behavior of the visitors and show the recommendations to the visitor at the bottom of the kiosk screen in the shopping cart, so that the visitor could add one or more of the recommended items to the order in a single tap. The key metric selected by the customer was  the gross margin percentage gained from items sold by recommendation $X$, i.e. gross margin of the items added from the recommender block $G_{rec}$ divided by total gross margin $G_{total}$ of the test segment during the selected timeframe (say, 1 day or 1 week):

\begin{displaymath}
  \ X= G_{rec}/G_{total}*100,
\end{displaymath}
where gross margin is calculated as
\begin{displaymath}
  \ G = R-C-T,
\end{displaymath}
and R is the  Revenue, C is the Cost of the goods sold, T is the Tax.

The customer has organized an online competition for a few teams developing recommender systems. The pool of models also included the customer's simple baseline model that implements several straightforward business rules such as "If the order contains a burger, then recommend a drink", "If the order contains a burger and a drink, then recommend french fries" etc. Thus we could freely compare, analyze and utilize in model training not only our historical data but the data of our competitors as well (the same was true for the other competing teams). We had access to the PostgreSQL database containing historical order data. At our disposal were: Order ID, session ID, restaurant ID, timestamp, and the set of purchased dishes.

\section{Related work}

The most common approach to building recommender systems is collaborative filtering (CF), first proposed, likely, by Goldberg et al. \cite{goldberg1992using}. It is based on discovering the items' or users' similarities from the user-item interaction matrix. See, for example, Su and Taghi \cite{su2009survey} for a survey of older works. CF is often accompanied by items and/or users features integration and matrix factorization techniques such as SVD, PCA, and others. See, for example, Polat and Wenliang \cite{polat2005svd} or Vozalis et al. \cite{vozalis2010collaborative}. 

Various works previously proposed CF for personal recommendations and reducing order time in the fast-food industry. For example, Azevedo and Wörndl \cite{azevedo2015adaptive}  suggested CF-inspired adaptive electronic menu for cafes and restaurants aimed to increase visitors' satisfaction and collect feedback. Chao et. al \cite{chao2016dish} have also used the skip-gram technique to retrieve dishes information from restaurant reviews. 

Maia and Ferreira \cite{maia2018context} enriched the CF-based food recommendation system by adding ingredients as features and contextual information such as location. The idea behind this work is an exploration of users’ preferences in conjunction with cultural, national features derived from users' location. Recent work by Gupta et al. \cite{gupta2021mood} suggested an integrated solution for cafes and restaurants. The user must enter their basic personal details so that the system could estimate his/her mood and make a personal recommendation based on their current mental condition.
 
The work by Wang et al. \cite{wang2020context} is likely the closest to ours in the terms of setting. In their Drive-thru recommendation service for Fast Food restaurants, the authors deal with session-based data and model it as a sequence of dishes added to the shopping cart. They utilize a transformer neural network to model dependencies related to the order of the dishes. It is noted the significance of the contextual data such as time, day of the week, location of the restaurant, etc., so the paper describes an extra transformer fully dedicated to the features of context.

Bonnin, Brun, and Boyer \cite{bonnin2008collaborative} have probably first suggested using a language model in a recommender system, although they did not go beyond working with Internet navigation artificial corpora. Valcarce et al. \cite{valcarce2015exploring, valcarce2016language} have also explored statistical language models for recommender systems, although the application area of these studies differs from described in this paper. Using a vector space with language models was first suggested by Valcarce et al. \cite{valcarce2017axiomatic} in a neighborhood-based recommender setting. Most recently, Zhang et al. \cite{zhang2021language} suggested the use of Pretrained Language Models, such as BERT, in recommender system, with limited success.

\section{The Dataset and Limitations}

\begin{figure}
  \includegraphics[width=0.45\textwidth]{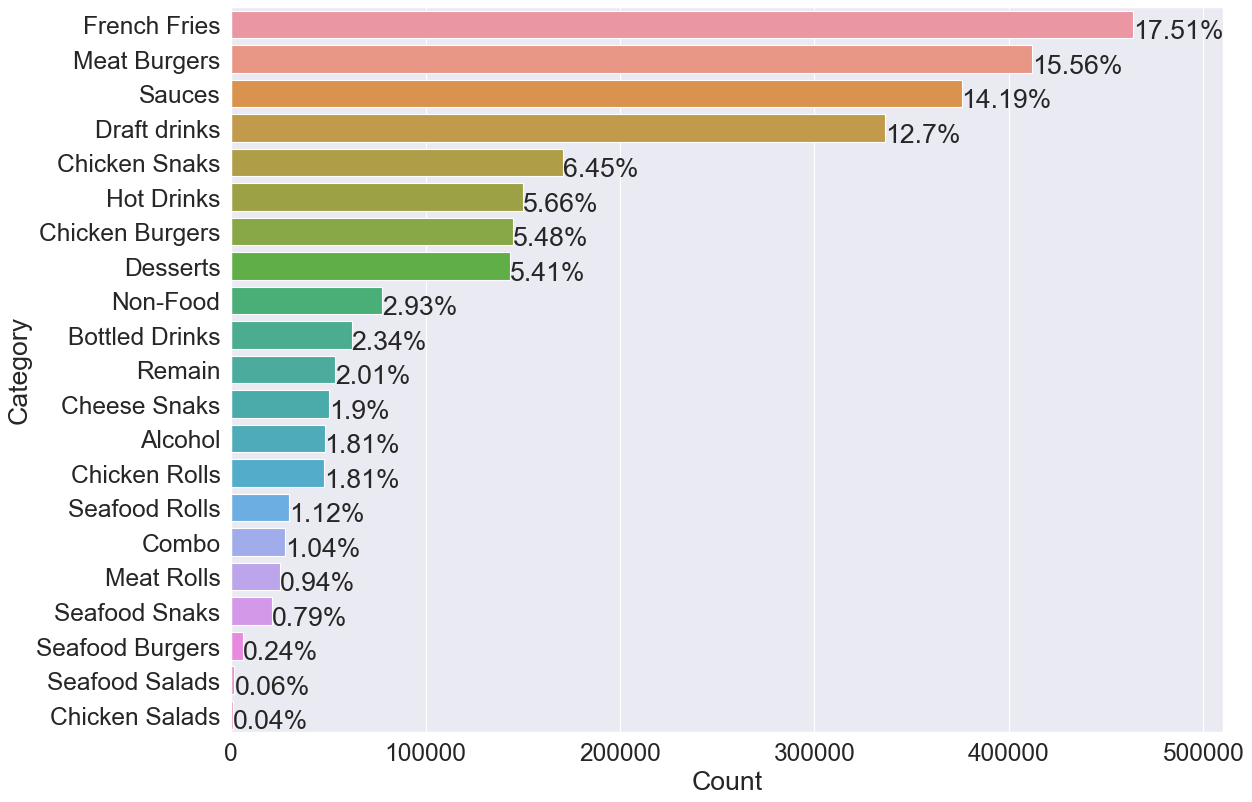}
  \caption{The distribution of all the purchases into categories.}
   \label{fig3}
\end{figure}

There are several peculiarities in the data we used that stem from the specific usage patterns of kiosks in a fast-food chain and result in a set of differences from the canonical recommender system datasets that often assume having historical personalized user-to-item interactions:
\begin{itemize}
\item no dish ratings are available, and all the feedback is completely implicit;
\item all orders are fully anonymous, thus personalized item-to-user recommendations are not feasible;
\item the number of items in the menu does not exceed 300, and there were a few dozen thousand orders per day, which results in a dense item-to-item matrix;
\item a flag pointing out that the item has been purchased by the recommendation was available.
\end{itemize}

Another aspect of the data is a one or two days delay between the moment a new dish goes live in the restaurants and the moment it becomes available in the database we work with. Thus we had to deal with "Out of vocabulary" (OOV) dishes during the inference. Initially, we replaced the OOV item with another one that is as close as possible according to the Normalized Levenshtein Distance Metric \cite{yujian2007normalized}. Later in this paper, we describe the final approach we used in production.

Let’s provide some exploratory analysis of the dataset. The customer’s database categorized all the items into three levels of categories. The order history available spanned several months.  Figure~\ref{fig3} shows the distribution of all the purchases into categories.

\begin{figure}
  \includegraphics[width=0.45\textwidth]{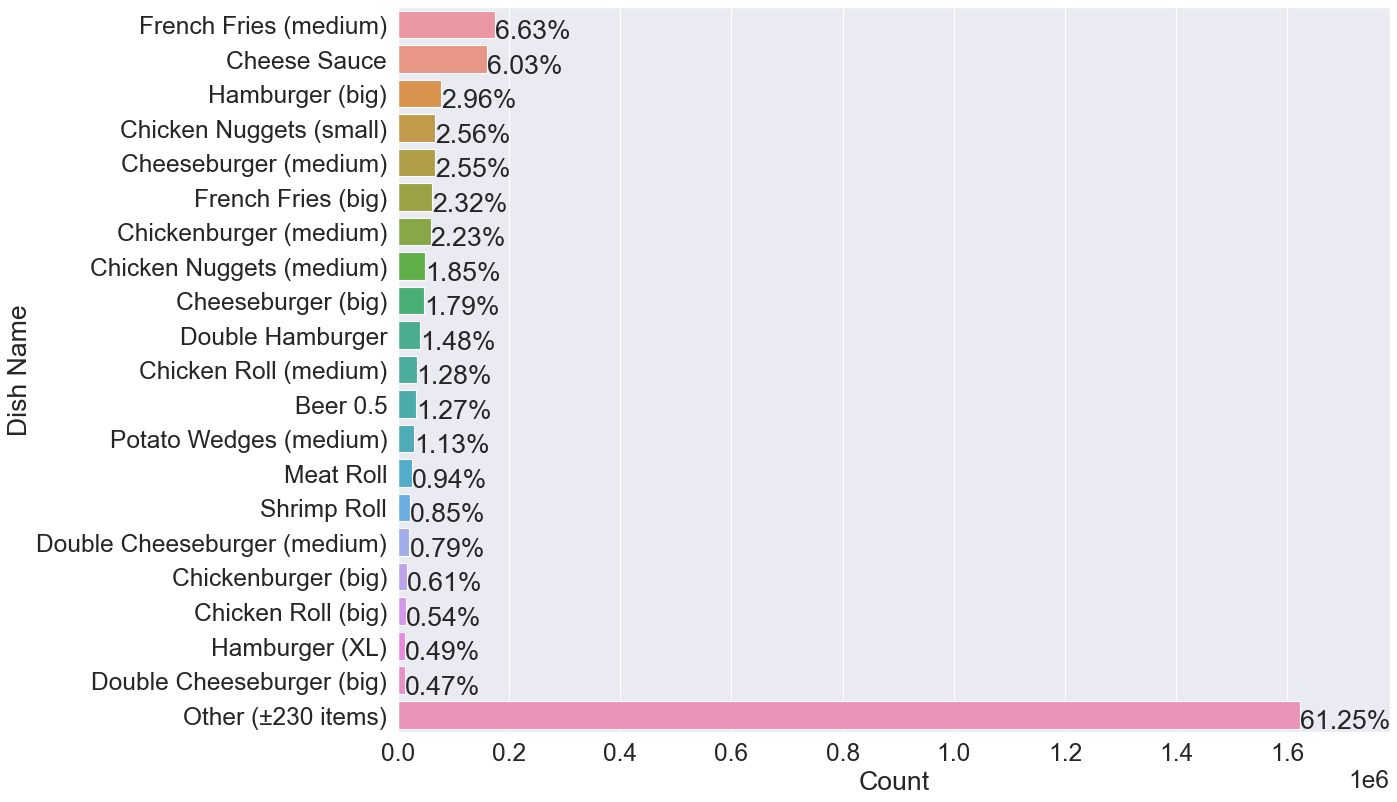}
  \caption{The distribution of dishes sold by the number of items.}
   \label{fig4}
\end{figure}

\begin{figure}
  \includegraphics[width=0.45\textwidth]{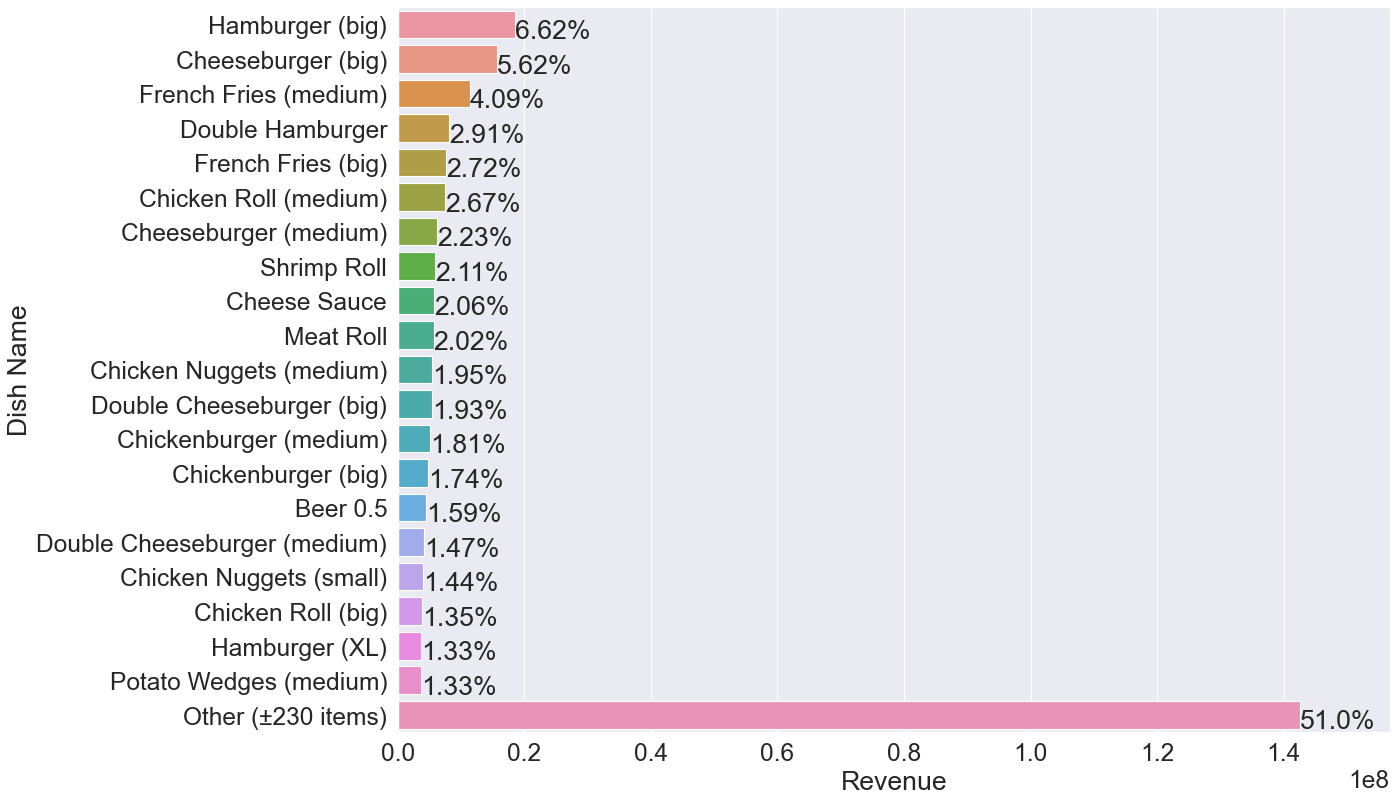}
  \caption{The distribution of the dishes sold by revenue.}
   \label{fig5}
\end{figure}

\begin{figure}
  \includegraphics[width=0.45\textwidth]{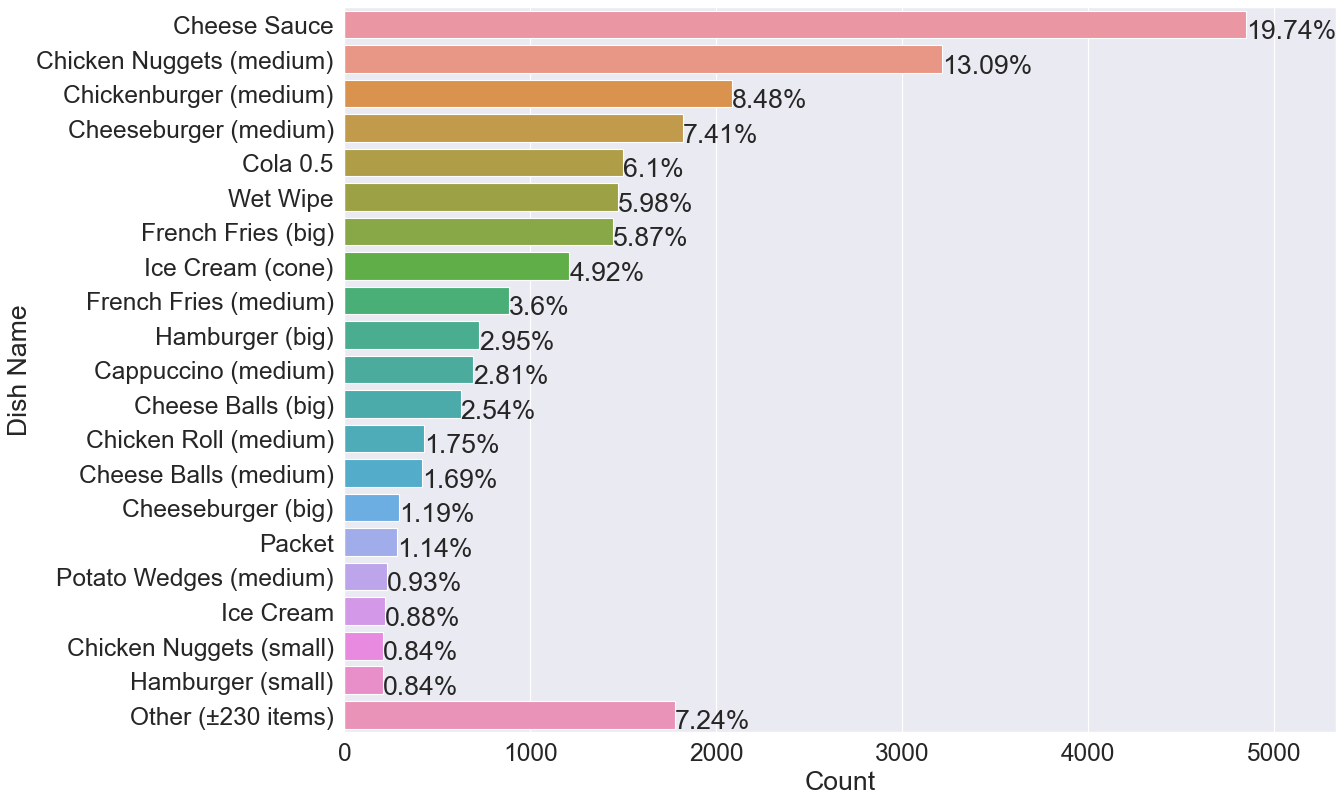}
  \caption{The distribution of recommended purchases by the number of items sold.}
   \label{fig6}
\end{figure}

\begin{figure}
  \includegraphics[width=0.45\textwidth]{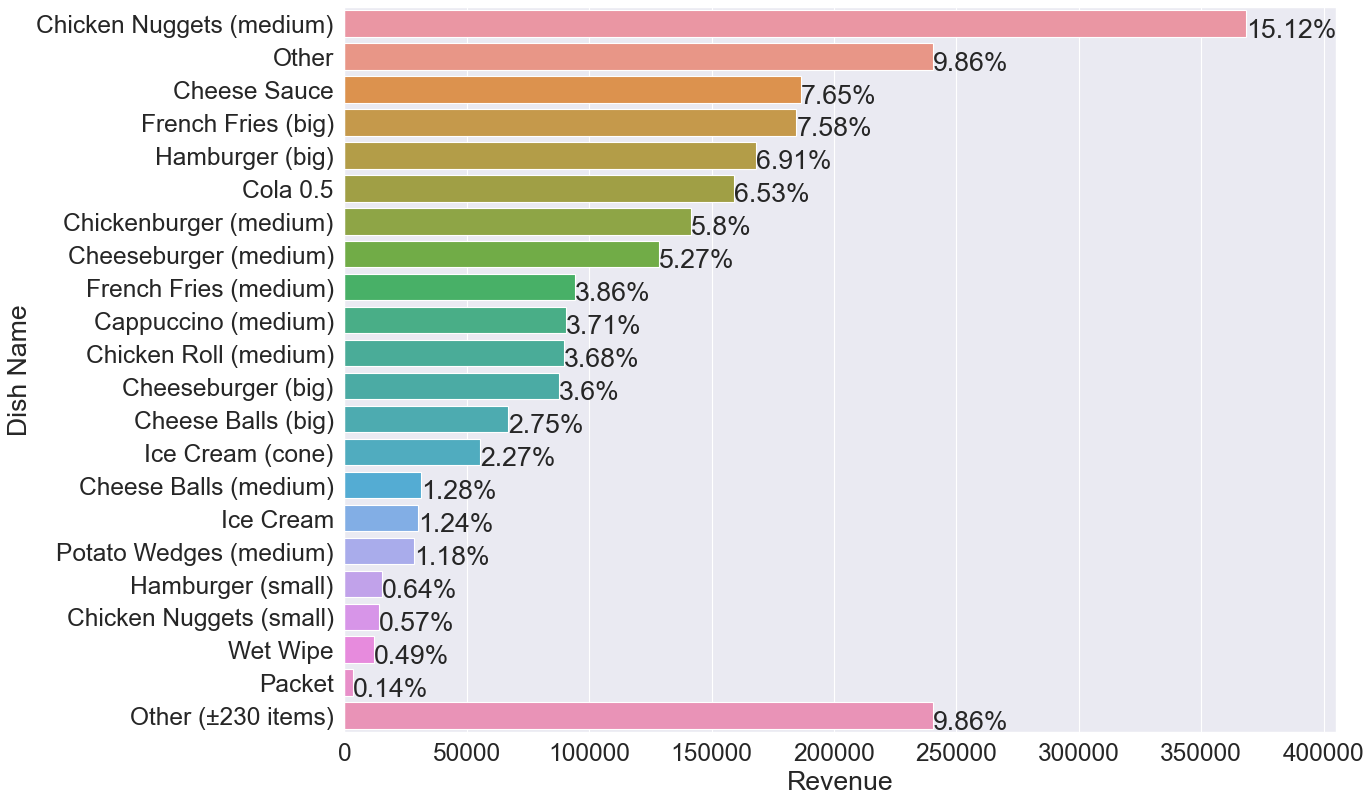}
  \caption{The distribution of recommended purchases by revenue.}
   \label{fig7}
\end{figure}

It is important to note that the top 20 items of the menu account for almost 40\% of the purchases by  number (Figure~\ref{fig4}) and almost 50\% of the purchases by revenue (Figure~\ref{fig5}). The top 20 items account for over 92\% of the items purchased from recommendations of the previous recommender systems. Figure~\ref{fig6} and Figure~\ref{fig7} show the distribution of the items purchased by recommendations in terms of their number and the revenue associated.

We can note that unlike the majority of recommendation datasets, a high density of the item-item matrix provides us with a sufficient amount of data on the one hand but suffers from redundancy and noise on the other one. The peculiarity of the fast-food restaurant is a significant skewness of the dish purchases' distribution. The major driver-items are burgers,  cold drinks and side dishes like french fries, thus the absolute majority of the orders contain items from the listed three groups. Hence, usage of classic collaborative filtering approaches leads to heavy biasing of recommendations to those items.

\section{Suggested approach}

We focused directly on increasing the gross margin percent and predicted the items relevant for the current cart. Because of the skewness of the distribution of fast-food chain visitors' preferences, based on the data provided above from the previous recommender systems, recommending just the top 8\% of the menu satisfies the needs of 90\% of visitors and brings 90\% of extra income. Considering that fact, we trained a model to perform the classification of the shopping cart into roughly 20 classes, each class representing an item to recommend, and recommended the top 4 classes predicted for the shopping cart.

\begin{figure}
  \includegraphics[width=0.45\textwidth]{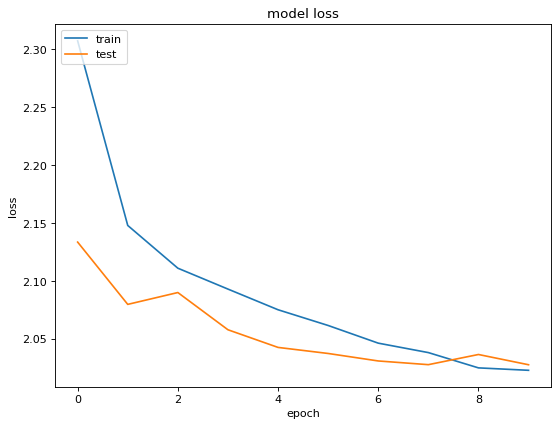}
  \caption{Model train and test losses.}
   \label{fig9}
\end{figure}

\begin{figure*}
  \includegraphics[width=0.9\textwidth]{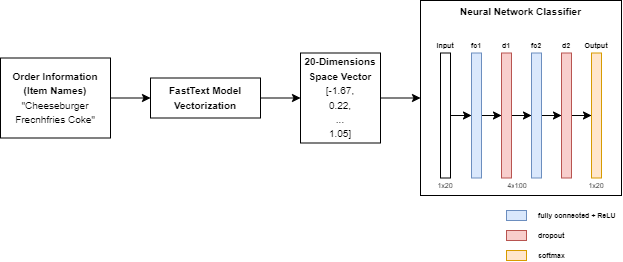}
  \caption{Model architecture.}
   \label{fig8}
\end{figure*}

To keep the models up-to-date we performed nightly training using the sliding fortnight data frame.
During the inference, each existing dish in the order contains the dish ID, dish name, and quantity, so there are two options to deal with them: lookup the dish metadata in the database by ID, or use dish name directly to "understand" the item. Even though the dish database lookup  seems to be a straightforward solution, it does not solve the problem of OOV and does not model inter-dish relationships, which might be useful for understanding the structure of the order. To tackle those issues we need a transformation of the dish name into vector space of fixed dimension, so that:
\begin{itemize}
\item semantically similar dishes like “hamburger” and “cheeseburger” are located closely, while semantically irrelevant ones - far away from each other;
\item same for behaviorally similar items, so that one could cluster together dishes of the main course, drinks, snacks, etc even though their names could be different like “brownie” and “cherry pie”;
\item we also need the transformation to be able to accurately estimate the "meaning" of previously unseen dishes (OOV) and properly locate them in the vector space.
\end{itemize}
FastText \cite{bojanowski2017enriching} fits well for the task  because it solves the OOV problem by splitting the previously unseen words into a set of N-grams and can be trained on a large amount of data in an unsupervised manner. FastText has also shown high efficiency in classification of short texts comparable to the shopping cart dish list \cite{zubchuk2021efficiency}.
Thus, the model we suggest contains two parts: a vectorizer that transforms the shopping cart into a vector in the vector space, and a classifier, operating with the vectors from the previous step. Training of each part is performed separately. First, we train the vectorizer in an unsupervised manner using all the available orders in the desired timeframe. Considering the relative consistency of the menu, we used 3 months timeframe. Second, we train a three-layer fully connected Neural Network Classifier. The classifier model had been trained with categorical cross-entropy loss \cite{zhang2018generalized} for 10 epochs. Model train and test losses during the training process are presented in Figure~\ref{fig9}.  The model structure is presented in Figure~\ref{fig8}.

\section{Results}

\begin{figure*}
  \includegraphics[width=0.9\textwidth]{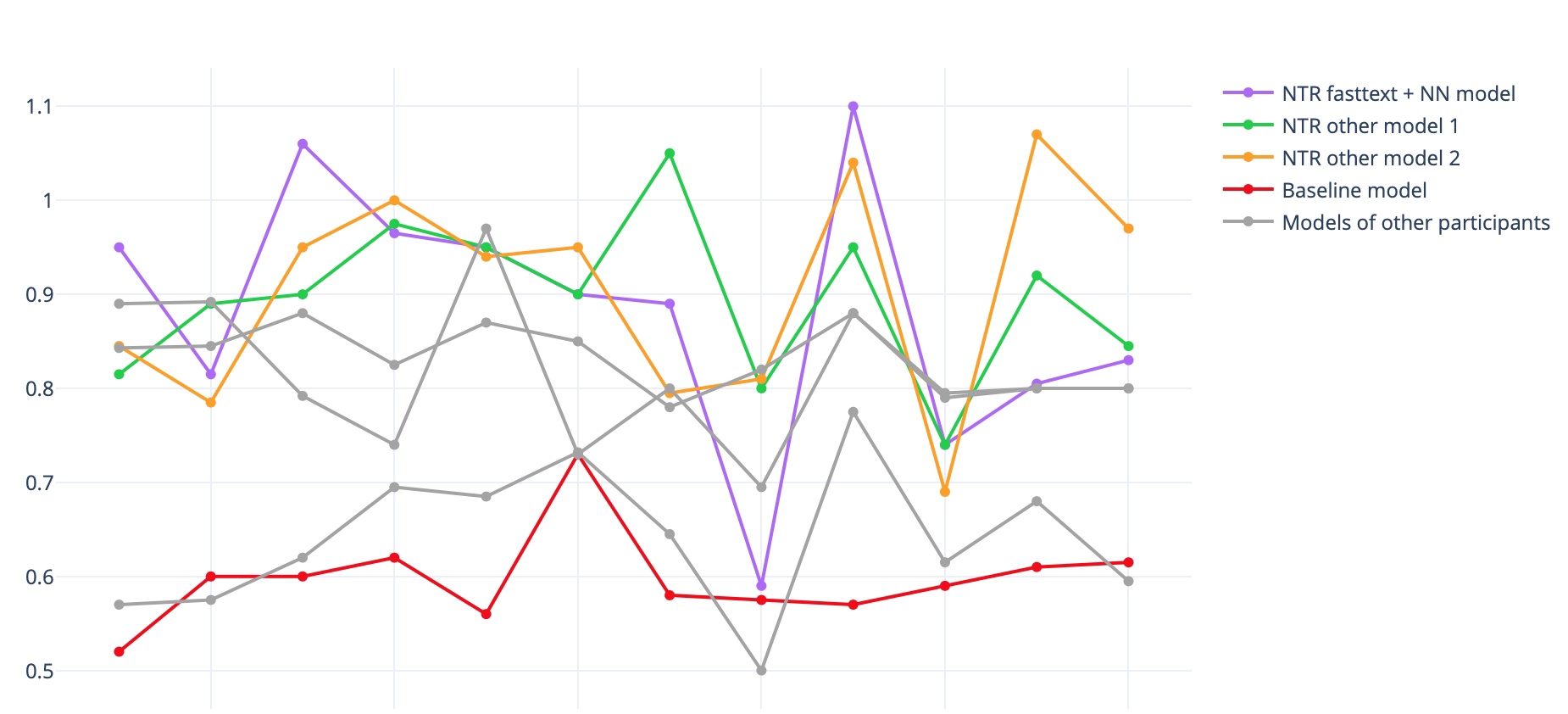}
  \caption{Timeline of percent of the added gross margin per day.}
   \label{fig10}
\end{figure*}

As the project was organized as a live A/B/C test, where several developer teams could compete in maximizing the percent of the added gross margin, we had an opportunity to compare our results with the results of other participants. Before going live we evaluated the models in the offline test.  We measured the ability to predict the recommended ground truth item, purchased by the user with Mean Average Precision at k (MAP@1 - MAP@4) \cite{he2018local}. We used a dataset of orders with the successful recommendations collected during a fortnight timeframe, i.e. all the orders contained the item which was recommended by any of recommender systems. We also used recommend percent metric, which is calculated as
\begin{displaymath}
  \ rec = O_g/O_a,
\end{displaymath}
where $O_g$ - orders where model guessed next item in top-4 predictions, $O_a$ - all orders.
Model metrics are presented in Table~\ref{tab:freq}.
\begin{table*}
  \caption{Offline metrics of recommender models}
  \label{tab:freq}
  \begin{tabular}{cccccl}
    \toprule
    Name&MAP@1&MAP@2&MAP@3&MAP@4&rec percent\\
    \midrule
    NTR FastText + NN Model&0.405&0.504&0.544&0.563&0.174\\
    NTR Other Model 1&0.285&0.365&0.399&0.418&0.128\\
    NTR Other Model 2&0.340&0.423&0.459&0.477&0.149\\
  \bottomrule
\end{tabular}
\end{table*}
Figure~\ref{fig10} demonstrates the behavior of the models in the live A/B/C test during a twelve days timeframe. As the number of models evaluated simultaneously was limited, we have replaced models in each slot from time to time. Thus, for the sake of consistency, we only provide comparative data for a limited timeframe.

The evaluation above shows that while the fastText model excels in the offline metrics and significantly outperforms the other models measured, in an online test it only outperformed the models of other competitors. The other models we have developed performed on par with the model described and even slightly better on average, although the difference is not statistically significant. Still, a different model with a more traditional architecture has been chosen for a production run.

The model described in this paper is represented in the diagram by the label ‘NTR fasttext + NN model’. ‘NTR other model 1' and 'NTR other model 2' are our models built on different principles not described here. As it is seen, all three demonstrate similar performance in the terms of extra gross margin percent, overcoming competitors. NTR and baseline models mean and standard deviation are presented in Table~\ref{tab:results}.

\begin{table}
  \caption{NTR and baseline models timeframes mean and standard deviation}
  \label{tab:results}
  \begin{tabular}{cccl}
    \toprule
    Name&Mean&Standard Deviation\\
    \midrule
    NTR FastText + NN Model&0.883&0.134\\
    NTR Other Model 1&0.895&0.081\\
    NTR Other Model 2&0.904&0.112\\
    Baseline Model&0.598&0.048\\
    Competitor 1 model&0.640&0.073\\
    Competitor 2 model&0.832&0.033\\
    Competitor 3 model&0.815&0.075\\
  \bottomrule
\end{tabular}
\end{table}

\section{Conclusions and future work}

The model suggested in this paper have exhibited good performance in real-life A/B/C tests, beating models from any other competitors. On the other hand, while the model presented here is significantly better than the other models we have studied in terms of offline metrics, the online metrics difference is not statistically significant. 

Despite the fact that the model beats competitors and demonstrates good performance, there still is room for improvement. Some user preferences may highly depend on the context, which is not considered in the model. Obviously, the majority of visitors prefer drinking coffee in the morning rather than in the evening. However, some restaurants are located on the highway, so coffee might be a good source of energy for drivers in the nighttime. The popularity of cold desserts such as ice cream or milkshakes decreases dramatically in cold time while consumption of alcoholic beverages depends on the day of the week, again, depending on the location of the restaurant. Figure 11 demonstrates the demand distribution for coffee (upper) and alcoholic beverages (lower) over time and day of week where 0 - Monday, 6 - Sunday. Red color stands for high demand and blue - for low.

\begin{figure}
  \includegraphics[width=0.45\textwidth]{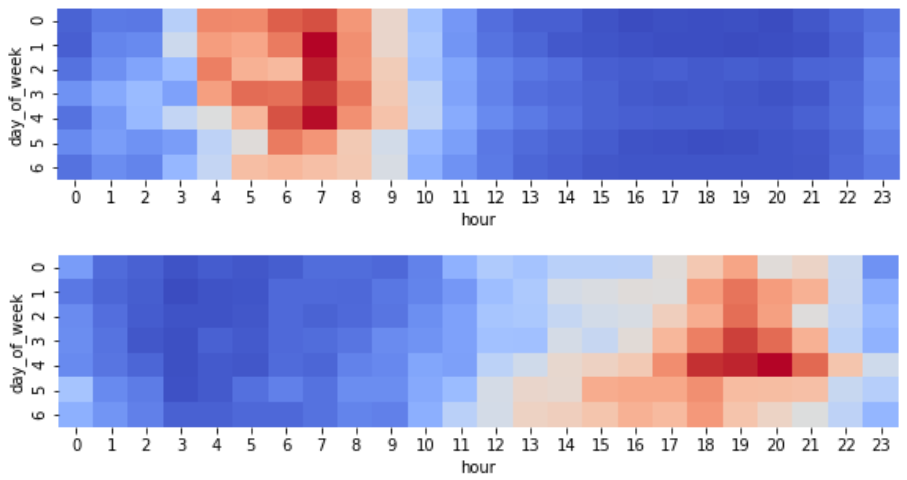}
  \caption{Distribution for coffee (upper) and alcoholic beverages (lower) consumption over the time of day and the day of week.}
   \label{fig11}
\end{figure}

There are many more dependencies that are not obvious but could be discovered in a latent manner in the process of machine learning. The model successfully discovers the dish features such as 'main course', 'drink', 'side dish' etc. in a latent way in the process of unsupervised training. However, feeding the item features to the model explicitly may also result in a better quality of the predictions. 
The basic data exploration demonstrates that the regular user follows a particular pattern while adding the dishes to the cart: he/she adds the main course such as a burger, a roll, etc., first, while desserts and snacks usually reside in the end. We do not exploit this pattern in our model so far, although it is potentially beneficial. All the above are directions of the further research and development.

\bibliographystyle{ACM-Reference-Format}
\bibliography{Languagemodelasrecommender}

%\appendix

\end{document}